\documentclass[sigconf]{acmart}

\AtBeginDocument{%
  \providecommand\BibTeX{{%
    \normalfont B\kern-0.5em{\scshape i\kern-0.25em b}\kern-0.8em\TeX}}}

\copyrightyear{2022} 
\acmYear{2022} 
\setcopyright{acmcopyright}\acmConference[CIKM '22]{Proceedings of the 31st ACM International Conference on Information and Knowledge Management}{October 17--21, 2022}{Atlanta, GA, USA}
\acmBooktitle{Proceedings of the 31st ACM International Conference on Information and Knowledge Management (CIKM '22), October 17--21, 2022, Atlanta, GA, USA}
\acmPrice{15.00}
\acmDOI{10.1145/3511808.3557122}
\acmISBN{978-1-4503-9236-5/22/10}
\usepackage{booktabs}
\usepackage{multirow}

\begin{document}


\title{Generating Persuasive Responses to Customer Reviews with Multi-Source Prior Knowledge in E-commerce}
\thanks{*Equal contribution}
\thanks{$\dag$Jiayi Liu is the corresponding author.}




\author{Bo Chen$^{*}$}
\affiliation{%
  \institution{Alibaba Group}
  \city{Hangzhou}
  \country{China}}
\email{herbert.cb@alibaba-inc.com}

\author{Jiayi Liu$^{*\dag}$}
\affiliation{%
  \institution{Alibaba Group}
  \city{Hangzhou}
  \country{China}
  }
\email{ljy269999@alibaba-inc.com}

\author{Mieradilijiang Maimaiti}
\affiliation{%
  \institution{Alibaba Group}
  \city{Hangzhou}
  \country{China}
}
\email{mieradilijiang.mea@alibaba-inc.com}

\author{Xing Gao}
\affiliation{%
  \institution{Alibaba Group}
  \city{Hangzhou}
  \country{China}
}
\email{gaoxing.gx@alibaba-inc.com}

\author{Ji Zhang}
\affiliation{%
  \institution{Alibaba Group}
  \city{Hangzhou}
  \country{China}
}
\email{zj122146@alibaba-inc.com}

\renewcommand{\shortauthors}{Bo Chen et al.}


\begin{abstract}
Customer reviews usually contain much information about one's online shopping experience. While positive reviews are beneficial to the stores, negative ones will largely influence consumers’ decision and may lead to a decline in sales. Therefore, it is of vital importance to carefully and persuasively reply to each negative review and minimize its disadvantageous effect. Recent studies consider leveraging generation models to help the sellers respond. However, this problem is not well-addressed as the reviews may contain multiple aspects of issues which should be resolved accordingly and persuasively. In this work, we propose a Multi-Source Multi-Aspect Attentive Generation model for persuasive response generation. Various sources of information are appropriately obtained and leveraged by the proposed model for generating more informative and persuasive responses. A multi-aspect attentive network is proposed to automatically attend to different aspects in a review and ensure most of the issues are tackled. Extensive experiments on two real-world datasets, demonstrate that our approach outperforms the state-of-the-art methods and online tests prove that our deployed system significantly enhances the efficiency of the stores' dealing with negative reviews.

\end{abstract}



\maketitle


\begin{figure}[!h]
    \small
  \centering
  \includegraphics[width=0.85\linewidth]{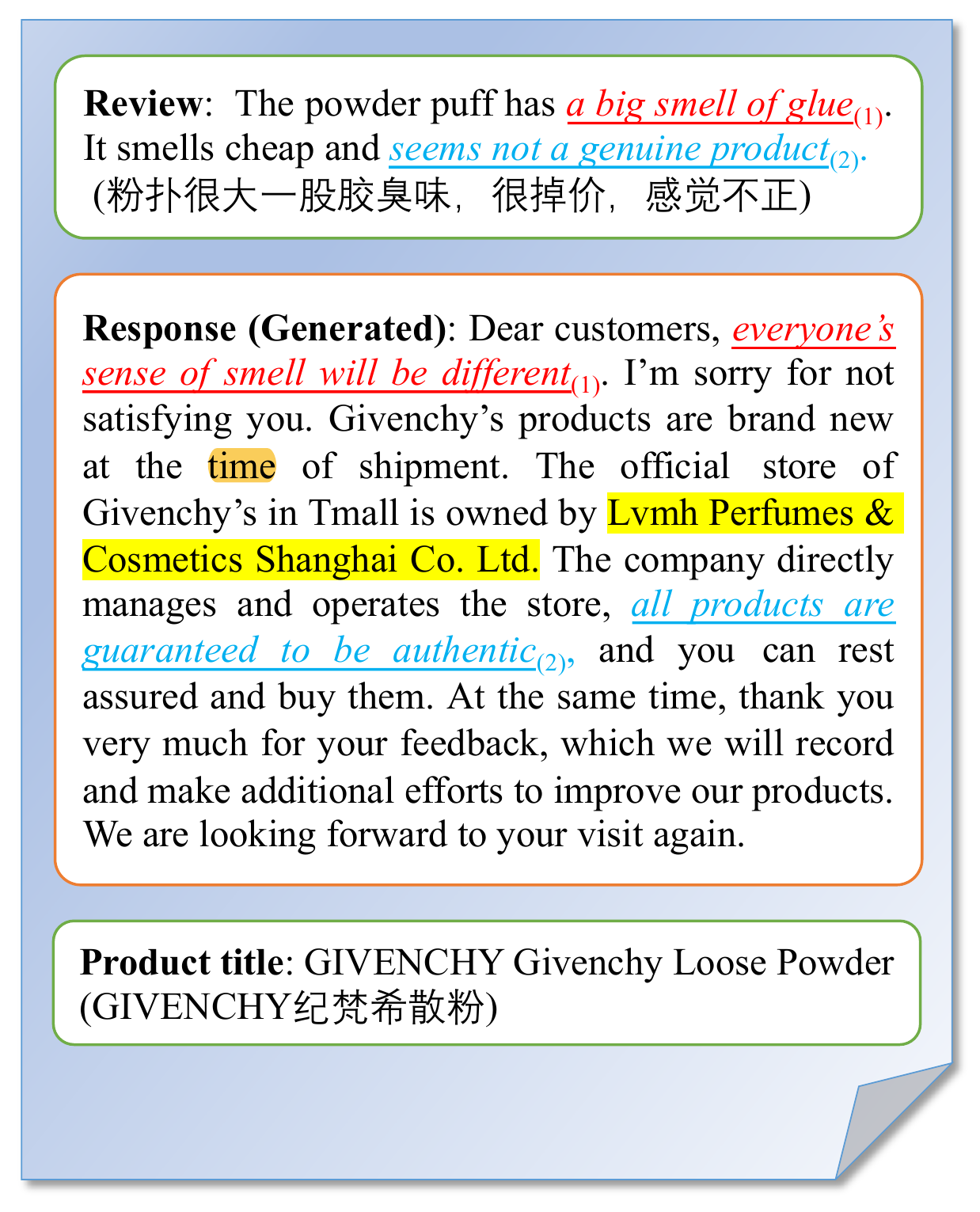}
  \caption{A case of the generated response. Different aspects of issue are marked with different color. Our model manages to deal with every mentioned aspect with persuasive and informative phrases. The company's name (highlighted in yellow) are copied precisely from our external knowledge. 
  }\Description{}\label{figure:sample}
\end{figure}

\section{Introduction}

With the development of the Internet, many users prefer to shopping online via various E-commerce platforms because it is more convenient for them to compare the prices or qualities of large varieties of goods. In this process, product reviews are of great importance for the buyers to take into account, as they offer real experiences from those who have already bought the product. 

While positive reviews are beneficial to the product seller, negative ones will largely influence other consumers and may lead to a decline in sales.
It has been proven that the presence of an organizational response to negative reviews is able to enhance potential customers' expectation regarding business trustworthiness and customer care~\cite{sparks2016responding}. %
Also, some statistical results in \cite{zhao2019review} demonstrate that sellers who provide high-quality and persuasive responses to the reviews tend to have higher sales volume. However, some unpersuasive responses such as ``Sorry for your bad experience. We will do better next time'' can neither help remove the customers' doubts, nor clear up the problems mentioned in the review. Therefore, it is of great importance for the sellers to carefully design some persuasive responses and minimize the disadvantageous influence of the negative reviews.

From our observation, a persuasive response needs to incorporate the following two characteristics:
\begin{itemize}
    \item \textbf{Reply meticulously to every mentioned issues.} Consumers care very much about whether all the problems mentioned in the negative review are properly solved. Failure to address some of them can be considered as tacit admissions of those problems. A fine case of persuasive responses is shown in Figure~\ref{figure:sample}, where the two dissatisfied aspects {\it (1) terrible smell} and {\it (2) fake product} are replied point by point;
    \item \textbf{Contain informative and detailed facts.} A high-quality and persuasive response should also provide extra product information or domain knowledge when needed. This can help convince the consumers of its good product quality and its authentication. For example, in Figure~\ref{figure:sample}, the response provides additional information that it is the official store of {\it Lvmh Perfumes \& Cosmetics Shanghai Co. Ltd} to prove its authenticity.
\end{itemize}

However, existing methods fail to generate review responses that satisfy the above characteristics well.
For example, \cite{zhao2019review} tries to address the problem of response generation for user reviews in the clothing domain, but the proposed method mainly relies on parametric memory~\cite{lewis2020retrieval}, which may leads to general and common responses. \cite{gao2019automating} proposes to integrate the meta information of app reviews, such as the category of app and the rating of review to improve the performance of generation. In the follow-up work~\cite{gao2020automating}, the authors make use of historical responses through retrieval, but they do not pay attention to tackling the multi-aspect problems. In this work, we propose a Multi-Source Multi-Aspect Attentive Generation model for persuasive response generation. Various sources of information are appropriately obtained and leveraged by the proposed model for generating more informative responses and promote seller's the brand image. A multi-aspect attentive network is proposed to automatically attend to different aspects in a review and meticulously reply every aspect. We conduct experiments on a publicly available dataset (i.e, Taobao Clothes) and a newly-constructed dataset  (i.e, Taobao Makeup). On automatic evaluation, our proposed method achieves state-of-the-art performance compared with baseline methods, bringing an absolute improvement of 7.58 and 6.02 BLEU score on Taobao Clothes and Taobao Makeup dataset, respectively. Human evaluation results demonstrate that our method can produce more persuasive responses than baseline methods. After this system is deployed online, we observed a significant enhancement for the efficiency of the stores' dealing with negative reviews.

In conclusion, our paper makes the following contributions:
\begin{itemize}
    \item We define the task of persuasive response generation and thoroughly analyze its challenges. To fairly evaluate the baseline methods on this task we also construct a novel dataset (i.e, Taobao Makeup), which will serve as a solid foundation for future research.
    \item We propose a novel Multi-Source Multi-Aspect Attentive Generation model to tackle this problem. The proposed model is able to adaptively encode various textual information and attend to different aspects of issues to generate persuasive and informative responses.
    \item We conduct extensive experiments on Taobao Clothes and Taobao Makeup dataset and the results demonstrate that our method outperforms baseline models on both automatic metric and human evaluation. Online deployment results shows that our method is useful in practice.
\end{itemize}

\section{Related Works}
\subsection{Seq2Seq Framework}
Text generation methods based on sequence to sequence framework have been widely studied. \cite{sutskever2014sequence} first introduced the Seq2Seq framework, which encodes the source sequence to a fixed length vector and decodes to generate the target sequence. The attention mechanism~\cite{bahdanau2014neural} is integrated into seq2seq framework to enhance the ability of capturing long term dependency, which is widely used in text generation tasks~\cite{xing2018hierarchical,liu2018table,kiddon2016globally}. The copy mechanism (Copynet ~\cite{DBLP:conf/acl/GuLLL16} and Pointer-Generator \cite{DBLP:conf/acl/SeeLM17}) is proposed to facilitate the model with the ability of copying words from source sequence, even works when there is an Out-of-Vocabulary (OOV) problem. \cite{DBLP:conf/nips/VaswaniSPUJGKP17} proposed transformer-based seq2seq framework based on multi-head attention mechanisms. Recently, pre-trained generative language models like  GPT-2~\cite{radford2019language}, UNILM~\cite{dong2019unified}, BART~\cite{lewis2020bart}, PALM~\cite{bi2020palm} have dominated this area. This pre-training and fine-tuning paradigm has been successful applied in many areas and is proved to have a significant impact on the downstream tasks by \cite{zhang2020pegasus,gururangan2020don}.

\subsection{Review Response Generation}

Review response generation task has been studied for different application domain in recent years. \cite{gao2019automating} proposes a Neural Machine Translation (NMT)-based neural network which encodes user reviews and generate developer's responses for mobile apps. Although some additional information (including the review's length, rating, predicted sentiment and etc) are adopted to better represent the semantics of the reviews, these source ignores the aspects of information from the developer and application itself, and thus maybe the responses are uninformative and unpersuasive. \cite{zhao2019review} proposes a model for the review response generation task in E-commerce platforms, which incorporates product information by a gated multi-source attention mechanism and a copy mechanism. However, \cite{zhao2019review} only use the product properties as external knowledge, which we argue is not enough to produce persuasive and informative responses. In our work we proposed to include more sources of prior knowledge and a better framework to model the intrinsic nature of this task.

\subsection{Retrieval-Augmented Generation}

Recently, multiple works \cite{lee2019latent,DBLP:conf/icml/GuuLTPC20,karpukhin2020dense} leverage pre-trained neural retriever to get knowledge from a large-scale textual corpus (e.g. Wikipedia).  \cite{DBLP:conf/eacl/IzacardG21} and \cite{DBLP:conf/nips/LewisPPPKGKLYR020} introduce dense retrieval method to augment generative models for open domain question answering, and achieve the state of the art performance. Our method also use a retriever to recall review-response pairs of similar reviews as textual knowledge. What is different is that, due to the limited number of retrieval pool and the specificity of our task, we use an ElasticSearch\cite{gormley2015elasticsearch} system with a simple ranking model rather than resource-consuming end2end dense retriever.



\section{Method}
\subsection{Task Definition}
Given a certain review $X = \{x_1, x_2,..., x_n \} $, where $n$ represents its length, our task is defined as generating an informative and persuasive response $Y = \{y_1, y_2,..., y_v \} $ to help solve the consumers' doubt and minimize the review's negative impact. Different from traditional response generation tasks, we need to guarantee certain factual information being conveyed correctly and persuasively, and thus we connect the item's id with its multiple sources of information, i.e, the 
item's title $T = \{t_1, t_2,..., t_u\}$, the item's product properties $P=\{(k_1,v_1),(k_2,v_2),...,(k_l,v_l)\}$ and the store's name $S = \{s_1, s_2,..., s_w\}$. Here $u$ and $w$ are the lengths of the corresponding word sequence, and the product properties are represented as $l$  key-value pairs. Finally, the model to solve this task can be formulated $f:(X,T,S,P)$->$Y$.

\subsection{Model Overview}

\begin{figure*}[h]
  \centering
  \includegraphics[scale=0.5]{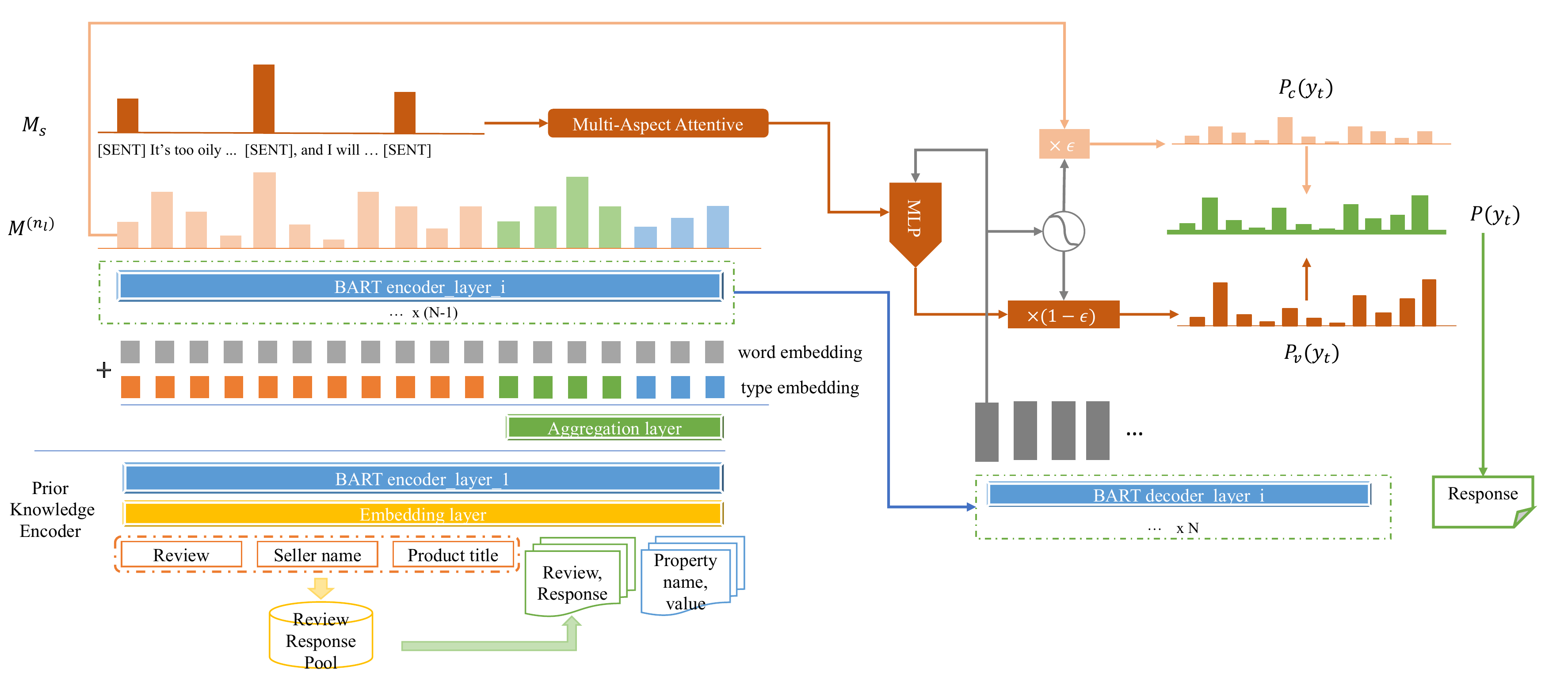}
  \caption{The Overview of the Whole Framework}\label{figure_overview}
\end{figure*}

Our proposed model is shown in Figure \ref{figure_overview}. The model has three main components, i.e., the multi-source prior knowledge encoder, the multi-aspect attentive network, and the pointer-generator network. The multi-source prior knowledge encoder first retrieves similar reviews from a pre-constructed review-response history pool via ElasticSearch \cite{gormley2015elasticsearch} and a ranking model. It then encodes them and fuses with existing encoded information (i.e.,the item's title, the item's product properties and the store's name) to form a memory matrix. Source dropout method is proposed to reduce the dependency on sole information and boost the robustness of the model. The multi-aspect attentive network is leveraged to help the model attend to different aspects of the review, not ignoring every mentioned issue. This is realized by pre-processing the reviews, segmenting them into different semantic parts and build a cross-attention network to attend to each part. Finally, the pointer-generator network guarantees the factual information, such as item title or store's name is properly incorporated, and then leverage the attended memory matrix to produce persuasive and informative responses.

\subsection{Multi-Source Prior Knowledge Encoder}
\subsubsection{Prior Knowledge Retrieval}\label{subsubsection:kgretrieve}
Despite the given review $X$, we connect its item id with several source of information including the item's title $T$, the item's product properties $P$ and the store's name $S$. However, we discovered that some obvious patterns should be learned from the historical review-response pairs, for example, some clarifying patterns, comforting words and reflection of feelings (which can be found in Figure~\ref{figure:sample}). Therefore, we propose to utilize ElasticSearch~\cite{gormley2015elasticsearch} and a ranking model to recall $m$ most relevant review-response pairs from a pre-constructed history pool, which is denoted as $R=\{(f_1,r_1),(f_2,r_2),...,(f_m,r_m)\}$. Here, $f_i$ is a recalled review, and $r_i$ is the corresponding human-written response to the review. Note that in our experiment we strictly split the train/dev/test set with no intersections on the reviews and responses, so that during training there is no risk of "information leaking" . To help the model not replying on certain source of information while ignoring others, we propose a new method named Source Dropout (SD) to boost the model's robustness. This is implemented by randomly discarding some information at a certain ratio and we found this method is simple but effective in our ablation study (section ~\ref{sec:ablation}), particularly when the recalled review-response pairs are absent.

\subsubsection{Text Knowledge Encoder}\label{subsubsection:textencoder}
Text knowledge encoder encodes the review $X$, the product title $T$, and the seller name $S$ into vectors. Formally, X,T,S is concatenated them into a long sequence $\hat{X} =\{x_1,x_2,...,x_n,[SEP],t_1,t_2,...,t_u,[SEP],s_1,s_2,...,s_w\}$, where the special [SEP] symbol marks the splitting position. The max number of $n,u,w$ are set to be fixed length, and we will truncate sentences when they are too long. We use the embedding layer and the first layer of BART encoder as the text knowledge encoder, which will convert $\hat{X}$ into embedded representations $\mathbf{\hat{X}}=\{\mathbf{x_1,x_2,...,x_{n+u+w+2}}\}$. Here, $\hat{\mathbf{X}}\in \mathbb{R}^{(n+u+w+2)\times D}$ and $D$ is the hidden size of BART encoder.

\subsubsection{Pair Knowledge Encoder} We propose to separately encode key-values in item properties $P$ and recalled review-response pairs $R$ as they contain multiple pair sequences. For item properties, we first concatenate a text pair with symbol [PAIR] as $p_i=\{k_i,[PAIR],v_i\}$, where $k_i$ and $v_i$ are key sequences and value sequences. Then $p_i$ is feed into the embedding layer to generate a matrix: $\mathbf{
p_i}$. Finally we use an aggregation function $g(\cdot)$ to convert the matrix $\mathbf{p_i}$ into property representation vectors, and we get $\hat{\mathbf{P}}\in\mathbb{R}^{l\times D}$ as encoded property features. In our experiment we use max-pooling function to reduce the amount of calculation:
\begin{equation}
  g(\hat{\mathbf{p_i}})=maxpool(\mathbf{p_i})
\end{equation}
Analogously, we get a matrix $\hat{\mathbf{R}}\in\mathbb{R}^{m\times D}$ to represent the recalled
$m$ review-response pairs.

\subsubsection{Multi-Source Fusion}
To distinguish different sources of information, we add a specific randomly-initialized type embedding vector to each output representation:
\begin{equation}
  \overline{\mathbf{X}}=\mathbf{\hat{X}}+ T(0),
  \overline{\mathbf{P}}=\mathbf{\hat{P}}+ T(1),
  \overline{\mathbf{R}}=\mathbf{\hat{R}}+ T(2),
\end{equation}
where $T(0),T(1),T(2)\in \mathbb{R}^D$ are the type embedding for input review, item properties and historical review-response pairs respectively.

Then we concatenate the three matrix to fuse multi-source information:
\begin{equation}
    \mathbf{M}=[\mathbf{\overline{X}:\overline{P}:\overline{R}}] \in \mathbb{R}^{(n+u+w+l+m+2)\times D}
\end{equation}
and feed $\mathbf{M}$ to the remaining BART layers. The output of the last layer is denoted as $\mathbf{M}^{(n_{l})}$, where $n_l$ is the number of BART encoder layers. 

\subsection{Multi-Aspect Attentive Network}

As is mentioned above, one review may contains multiple aspects of issue, and it is of vital importance to take care all of them meticulously or the consumers should not be satisfied. In this part we provide a pre-processing method and build multi-aspect attentive network to attend to different parts of the review when decoding.

Specifically, we pre-process each review by segmenting them into different semantic parts. This is done by heuristically separated the sentence based on some punctuation marks such as a comma, a full stop, and a exclamation mark, as we assume that these certain punctuation represent a pause within a sentence. Meanwhile a special token [SENT] is inserted between each semantic fragment. For instance, a review like \textit{"It’s oily after putting it on for at most two hours, and it smells weird."} will be re-constructed into \textit{"[SENT] It’s oily after putting it on for at most two hours\textbf{,} [SENT] and it smells weird\textbf{.} "}. Other semantic segmentation methods can be explored in future work.

The indexes of multiple inserted [SENT] token is denoted as  $I_s=\{s_1,s_2...,s_m\}$ and their  corresponding aspect-level representations are gathered by:
\begin{equation}
    M_s=[(h_{s_1}^e)^T:(h_{s_2}^e)^T:...:(h_{s_m}^e)^T],s_i \in I_s 
\end{equation}
Here $h_i^e$ is the i-th row vector of encoder output matrix $M^{(n_{l})}$ and $T$ means matrix transposition. During the decoding process, the hidden state of current decoder step $h_t$ should attend to these different semantic aspects, and produce a importance weight distribution:
\begin{equation}
    a_t=\sum\limits_{s_i \in I_s} \lambda_{t,i}*h_{s_i}^e,
\end{equation}
where $\lambda_{t,i}$ is the normalized importance weight over semantic parts. It is calculated by:
\begin{equation}
    \lambda_{t,i}=\frac{exp(e_{t,i})}{\sum\limits_{s_k \in I_s}exp(e_{t,s_k})},
\end{equation}
and $e_{t,i}$ is the attention energy vector for aspect $h^e_{s_i}$:
\begin{equation}
    e_{t,i}=w_1^T*tanh(W_1*(h^e_{s_i})^T+W_2*h_t+b_1).
\end{equation}
With this multi-aspect attentive network, the model learns to target at every important issue when generating responses. With a two layer perceptron network, the probability of words are produced based on
the current decoding states $h_t$ and multi-aspect attended vector $a_t$:
\begin{equation}
    P_v(y_t)=softmax(W_5*(W_3*h_t+W_4*a_t+b_2)+b_3),
\end{equation}
where, $W_1,W_2,W_3,W_4 \in \mathbb{R}^{D \times D},b_1,b_2 \in \mathbb{R}^{D\times 1}, W_5 \in \times \mathbb{R}^{D\times |V|},b_3 \in \mathbb{R}^{1\times |V|}$ are learnable parameters.

\subsection{Pointer-Generator Network}
Some factual information should be accurately incorporated into the generated response, such as the product title and the store's brand names. And thus, pointer-generator network \cite{DBLP:conf/acl/SeeLM17} fits this situation when the responses need to copy certain tokens from the multi-source input. In particular, we derive the copied tokens from the output of text knowledge $\hat{X}$ because of their frequently occurrence in responses.

We take the first $l_x=n+w+u+2$ columns of $M^{(n_{l})}$ as the representation of $\hat{X}$, 
namely,
\begin{equation}
    X^{n_l}=M^{(n_{l})}[0:l_x],
\end{equation}
here, $X^{n_l}$ is the representation of $\hat{X}$. While decoding, the pointer generator will first calculate a attention score $\beta_{t}$ over $X^{n_l}$ and aforementioned hidden state $h_t$:
\begin{equation}
    \beta_{t,i}=\frac{exp(\alpha_{t,i})}{\sum\limits_{l_x}exp(\alpha_{t,l_x})},
\end{equation}
where $\alpha_{t,i}$ comes from:
\begin{equation}
\alpha_{t,i}= w_6^T*tanh(W_6*(X^{n_l}_i)^T+W_7*h_t+b_6).
\end{equation}
Here, $W_6,W_7 \in \mathbb{R}^{D \times D},b_6 \in \mathbb{R}^{D\times 1}, W_5 \in \times \mathbb{R}^{D\times |V|}$ are learnable parameters.

As a result, $P_c(y_t)$ is the copying probability over the vocabulary extended with tokens in $\hat{X}$, formulated as 
\begin{equation}
    P_c(y_t)=\sum_{i:x_i=y_t}\beta_{t,i}
\end{equation}
The final probability $P(y_t)$ of generating word $y_t$ is defined as a mixture of the extended
vocabulary distribution $P_v(y_t)$ and the copy distribution $P_c(y_t)$:
\begin{equation}
    P(y_t)=\epsilon*P_v(y_t)+(1-\epsilon)*P_c(y_t)
\end{equation}
\begin{equation}
    \epsilon=sigmoid(w_{\epsilon}^T*h_t+b_{\epsilon})
\end{equation}
Here, $b_{\epsilon} \in \mathbb{R},w_{\epsilon} \in \mathbb{R}^{D \times 1}$ are learnable parameters.
We optimize the whole model with cross-entropy loss function:
\begin{equation}
    L(\theta;(\mathcal{X},\mathcal{Y}))=-\sum\limits_{(X,T,S,P,R) \in \mathcal{X},
    Y \in \mathcal{Y}}\sum\limits_{y_t \in Y}log(P(y_t))
\end{equation}

\section{Dataset Construction}

\subsection{Acquisition of Data}\label{section:da}
\subsubsection{Taobao Clothes}
The Taobao Clothes dataset, proposed by \cite{zhao2019review}, is widely used in this task. It contains 100k (review, product information, response) triples which all belong to the Clothes category. Our attempt to link each triple with its product title failed because the corresponding logs are expired. 

\subsubsection{Taobao Makeup}
To examine the generalization performance over different domains, we gather another 121k reviews and corresponding information from taobao.com where we construct the Taobao Makeup dataset. Makeup is chosen as our target category as it contains abundant (100+) sub-categories vary from cleanser, face powder to lipstick. To make the task more challenging and practical, we use a off-the-shelf sentiment analysis tool to filter out positive reviews and keep negative ones. The item titles and product properties are collected from its online detail page. We further built an ElasticSearch index based on the training set. Each retrieve query requires an exact match of the store's name and sub-category id, and an approximate match of the review and item title. We care more about the review by assigning a boost weight of 2.0 for review and 1.0 for item title. After the retrieving we use a ranking model to choose top $m$ similar data pairs.

\begin{table*}[!htbp]
\caption{Details of the Taobao Clothes and Taobao Makeup datasets.}\label{table:data}
\begin{tabular}{@{}ll|cc@{}}
\toprule
 &                     & Taobao Clothes       & Taobao Makeup        \\ 
\midrule
\multirow{3}{*}{number of}                                                                           & \multicolumn{1}{l|}{train}             & 80000                & 104532               \\
& \multicolumn{1}{l|}{valid}             & 10000                & 2986                 \\
& \multicolumn{1}{l|}{test}              & 10000                & 14054                \\ 
\midrule
\multirow{3}{*}{\begin{tabular}[c]{@{}l@{}} length of  \\ (min/avg/max)\end{tabular}} 
 & \multicolumn{1}{l|}{title}            & -                    & 4/33.1/61            \\
 & \multicolumn{1}{l|}{review}       & 19/61.6/360          & 1/32.0/665           \\ 
 & \multicolumn{1}{l|}{response}    & 23/117.8/415         & 18/131.1/562         \\
\midrule
\multirow{2}{*}{\begin{tabular}[c]{@{}l@{}}number of\\ (min/avg/max)\end{tabular}} & \multicolumn{1}{l|}{\multirow{2}{*}{properties}} & \multirow{2}{*}{1/14.2/26} & \multirow{2}{*}{1/10.6/21} \\  & \multicolumn{1}{c|}{}   &       &         \\
\midrule
\multirow{3}{*}{\begin{tabular}[c]{@{}l@{}} relevance of  \\ (r/p/f1)\end{tabular}}                    & \multicolumn{1}{l|}{title}             & -                    & 0.0443/0.1541/0.0668 \\
 & \multicolumn{1}{l|}{properties}          & 0.0204/0.0233/0.0204 & 0.0589/0.0648/0.0561 \\
 & \multicolumn{1}{l|}{recalled responses}   & -                    & 0.7955/0.3261/0.4375 \\ 
\bottomrule
\end{tabular}
\end{table*}

\subsection{Dataset Statistics}
The detailed statistics of the Taobao Clothes and Taobao Makeup dataset is shown in the table \ref{table:data}. The train/valid/test split of the proposed Taobao Makeup dataset is mainly based on the principle that no intersection of reviews and responding between different splits. We sample more data for test set to increase the challenge. The minimum/average/maximum length of the title/response/review is also shown in this table.

For better understanding which kind of source contributes most to generating a response, we calculated the relevance score between the target response with title, properties and recalled responses using the following word-level criteria:
\begin{equation}
\begin{aligned}
    r_i=&\frac{|H_i \cap R_i|}{|R_i|}\\
    p_i=&\frac{|H_i \cap R_i|}{|H_i|}\\
    f1_i=&\frac{2*p_i*r_i}{p_i+r_i}\\
\end{aligned}
\end{equation}
Here $i$ means the i-th sample of the data, $H_i$ is the word set of different source such as title or property, and $R_i$ is the word set of the i-th target response. We tokenized the word with Jieba\footnote{https://github.com/fxsjy/jieba}. As is shown in Table \ref{table:data}, the relevance scores of recalled responses significantly surpass other sources, from which the model can learn many responding verbal tricks or patterns. However, we do not desire the model to fully rely on this source as the item title and review itself should much be considered, and thus Source Dropout method is proposed to effectively deal with this problem. The title relevance $r$ is much more higher than properties, as expert-written response often contains its product name. 

\section{Experiments}

\subsection{Evaluation Metrics}
\subsubsection{Automatic Evaluation Metrics}
We follow \cite{zhao2019review} to use BLEU and ROUGE as the evaluation metrics for estimating generation accuracy. In addition, the distinct~\cite{li2016diversity} score is calculated in our experiments to examine the word-level diversity of responses. Details of these metrics are as follows:
\begin{itemize}
\item {BLEU}: BLEU is widely used in text generation task and it measures word-level overlap between the generated sentence and the ground-truth. As is done in \cite{zhao2019review}, we calculate the BLEU score with the off-the-shelf NLTK toolkit\footnote{https://www.nltk.org}. 
\item {ROUGE}: ROUGE is also a widely-used word-level evaluation metric. We report ROUGE-1, ROUGE-2 and ROUGE-L in this paper.
\item {Distinct}: The Distinct score is first proposed by \cite{li2016diversity}, which is to measure the n-gram diversity a corpus or a sentence. Note that we do not follow \cite{zhao2019review} to calculate the Distinct score at the corpus level. We use sentence-level Distinct score to reflect more information on each generated response.
\end{itemize}
\subsubsection{Human Evaluation Metric}
To evaluate the persuasiveness of different methods, we also conduct human evaluation in our experiments. We randomly sampled 500 cases and invited three annotators to evaluate the generated responses
of different models. For each sample (review, seller name,product title,response), annotators were asked to give a score from {1,2,3,4,5}, which complies with the following criteria:
\begin{itemize}
    \item 1 means
the response is irrelevant and ungrammatical;
\item 2 means the response is fluent and grammatically correct, but it's irrelevant to the review;
\item 3 means that the response is fluent and grammatically correct, and is related to the review, but lacks persuasiveness. This suggests a safe and general reply.
\item 4 indicates that the response is smooth, relevant to review and persuasive, but missing some part of aspect mentioned in the review.
\item 5 indicates that the response is not only fluent, relevant to the review, but also fully and professionally replies each aspect of the review.
\end{itemize} 
\subsection{Baseline Methods}
We  compare our proposed method with following baselines:
\begin{itemize}
    \item {\textbf{EPI}}: EPI~\cite{zhao2019review} is the state of the art E-commerce review response generation method which uses an additional encoder for product properties, we get the result with their code.
    \item {\textbf{BART (encyclopedia)}}: We use Baidu Encyclopedia data (9.6GB, 9.58 million sentences) to pre-train a BART model following existing pre-training scheme, which treats the transformer network as a sequence to sequence de-noising auto-encoder. For fine-tuning, we concatenate the input review, the item properties and other different sources separated with a [SEP] symbol as the input. Between each property's key and value a [PAIR] symbol is added to reflect their relationship. Then the model is fine-tuned on the target datasets in a traditional seq2seq fashion.
    \item {\textbf{BART (review)}}: We use (review, title, seller name, response) quadruples to pre-train the BART model. The corpus contains 8.35 million quadruples and is collected with no overlap with the valid/test set in Taobao Clothes and Taobao Makeup. We add a type embedding for each source in the embedding layer and concatenate the input like BART (encyclopedia). After pre-training, we fine-tune the model on the target datasets. The only difference between BART (encyclopedia) and BART (review) is their pre-training methods.
    \item {\textbf{MsMAAG}}: The proposed \textbf{M}ulti-\textbf{s}ource \textbf{M}ulti-\textbf{A}spect \textbf{A}ttentive \textbf{G}eneration model in this work. For a variant of this model, the multi-aspect attentive network is removed and we denote this model (\textbf{MsG}). Note that for Taobao Clothes dataset, we don't have the recalled review-response pairs $P$ and product title $T$.
\end{itemize}
\subsection{Hyper Parameters}
For the EPI model, we use the same settings as reported in \cite{zhao2019review}. For all BART-based models, we set the hidden size as $D=768$ and transformer layer num as $n_l=6$. The warm up\cite{he2016deep} rate is $\gamma=0.075$, the batch size is $B=20$ and the number of beams in decoding progress is $n_B=3$. The learning rate for BART (encyclopedia) and BART (review) is set to $\delta=5e-5$, while for MsMAAG and MsG model is $\delta=1e-4$ to achieve their best performance. Models are trained for 10 epoches using the Adam optimizer, and teacher forcing training paradigm\cite{williams1989learning} is used for faster convergence. Through hyper-parameter tuning we found the best number of the recalled review-response pairs is $m=19$. Other parameter of BART model can be found in \cite{lewis2020bart}.

\begin{table*}[]
\caption{Automatic evaluation results on BLEU, ROUGE-1/2/L (r-1/2/l) and Distinct (Dist-1/2/3/4). }\label{table:automatic_eval}
\begin{tabular}{@{}lcccccccccc@{}}
\toprule
                                    & \multicolumn{5}{c}{Taobao Clothes}                                                            & \multicolumn{5}{c}{Taobao Makeup}                                                                                                                           \\ \cmidrule(l){2-11} 
                                    & BLEU           & r-1            & r-2            & r-l            & Dist-1/2/3/4             & BLEU                      & r-1                       & r-2                       & r-l                       & Dist-1/2/3/4                                 \\ \midrule
\multicolumn{1}{l|}{EPI}            & 15.61          & 36.71          & 18.07          & 28.84          & 0.623/0.809/0.848/0.856 & 12.42                     & 30.39                     & 15.36                     & 27.86                     & 0.494/0.624/0.658/0.664                     \\ 
\multicolumn{1}{l|}{BART (encyclopedia)}    & 12.40          & 34.89          & 14.77          & 24.95          & 0.683/0.887/0.932/0.940 & 28.29                     & 45.61                     & 29.35                     & 39.29                     & 0.692/0.888/0.927/0.935                     \\
\multicolumn{1}{l|}{BART (review)}   & 16.86          & 38.64          & 18.98          & 28.68          & 0.691/0.898/0.939/0.944 & 33.62                     & 50.66                     & 35.39                     & 44.83                     & 0.705/0.905/0.941/0.945                     \\
\midrule
\multicolumn{1}{l|}{MsG (ours)}            & 21.75          & 41.77          & 23.28          & 32.95          & 0.680/0.885/0.926/0.931 & 39.15                     & 54.83                     & 40.60                     & 49.49                     & 0.709/0.911/0.947/0.951                     \\
\multicolumn{1}{l|}{MsMAAG (ours)}        & \textbf{24.44} & \textbf{43.64} & \textbf{25.67} & \textbf{34.99} & 0.688/0.895/0.936/0.940 & \textbf{39.64}            & \textbf{55.32}            & \textbf{41.29}            & \textbf{50.11}            & 0.707/0.909/0.945/0.949                    \\ \bottomrule
\end{tabular}
\end{table*}
\subsection{Automatic Evaluation Results}

\begin{table}[]
\caption{Human evaluation results on persuasiveness. }\label{table:human}
\begin{tabular}{@{}lcccccc@{}}
\toprule
         & 1       & 2      & 3       & 4      & 5       & $score_{avg}$ \\ \midrule
EPI      & 66.60\% & 7.02\% & 20.58\% & 1.15\% & 4.66\%  & 1.71          \\
BART & 5.36\%  & 12.30\% & 30.33\% & 9.26\% & 42.75\% & 3.71          \\ \midrule
MsG      & 4.85\%  & 13.09\% & 27.27\% & 10.06\% & 44.73\% & 3.76         \\
MsMAAG  & 2.88\%	&11.38\%	&32.63\% &	8.13\% &	45.00\% &3.81          \\ \midrule
Human    & 5.84\%  & 6.04\% & 28.14\% & 4.64\% & 55.34\% & 3.95          \\ \bottomrule
\end{tabular}
\end{table}

The automatic evaluation results are shown in Table \ref{table:automatic_eval}. For BELU and ROUGE metrics, We can conclude that our proposed MsMAAG model (multi-source encoder, multi-aspect attentive network and pointer generator) is effective and achieves the state-of-the-art performance on both datasets. The comparison between MsG and MsMAAG models indicates that our proposed multi-aspect attentive network yield better performance, with an 2.69/0.49 points BLEU increment on Taobao Clothes/Taobao Makeup dataset, respectively. The enhancement on Taobao Clothes is bigger than on Taobao Makeup dataset, as our observation suggests the reviews on Taobao Clothes having more numbers of aspects and longer length (as is shown in Table~\ref{table:data}). This trend is also valid for ROUGE metric. For Distinct metrics, our MsMAAG model ranks second on Taobao Clothes, and first on Taobao Makeup, demonstrating some ability of producing diverse responses. Additionally, we find that the pre-trained BART model is generally better than the EPI method, but on the Taobao Clothes dataset, the BART model pre-trained on the Baidu Encyclopedia corpus is worse than the EPI model. This may due to the fact that Taobao Clothes dataset is more domain-specific than general Taobao Makeup dataset which includes more than 100 subcategories. For Distinct score, we can summarize that BART-based model generate more diverse results than EPI, and we discover that EPI sometimes produce some unreadable repetitive sentences. This is also proved in our human evaluation part.



\subsection{Human Evaluation Results}

The results of human evaluation on persuasiveness is shown in Table \ref{table:human}, where we report the
percentage of each score on Taobao Makeup dataset. The averaged score is given in the last column of the table. From this table we observe that our proposed MsMAAG model outperforms others in terms of persuasiveness. Comparing MsG and MsMAAG models, a 1.33\% increase in avaraged score is realized through the multi-aspect attentive network. We can see that with this network the portion of score 1,2,4 diminish and the they have been promoted to score 3,5, which means more related and thorough responses are generated. The EPI model tend to produce repetitive entities and thus some of them are ungrammatical and 66.6\% of the responses are scored to 1. This phenomenon can be observed in the Case Study section. Comparing BART model with MsG model, a 1.34\% increase due to the multi-source encoder and pointer-generator network. Note that when evaluating BART model, annotators find some hallucination problem as this model is prone to find product titles/store names from the whole corpus. This can cause some problems when the model is deployed online. Finally, we report the human results for reference. The percentage of score 5 with regard to human results are significantly higher than all the models, indicating that there is much room for optimization of these models. Intriguingly, the percentage of score 1 of BART, MsG, MsMAAG model is lower than the human results, which means that large-scale pre-trained corpus contributes to some self-correction ability in grammar.

\subsection{Ablation Study} \label{sec:ablation}
As in previous sections we already analyze the contribution of Multi-aspect attentive network, in this section we only examine the effect of Multi-source prior knowledge and Source Dropout (SD) method.

\begin{table}[]
\caption{Ablation study on Taobao Makeup dataset.}\label{tab:remove_source}
\begin{tabular}{@{}lllll@{}}
\toprule
                              & BLEU  & r-1   & r-2   & r-l   \\ \midrule
MsMAAG Model                  & 39.64 & 55.32 & 41.29 & 50.11 \\
\quad-w/o properties          & 39.40 & 55.15    & 41.07     & 49.90 \\
\quad-w/o recalled pairs      & 38.01 & 53.82    & 39.54     & 48.34 \\ \midrule
\quad-w/o source dropout (SD) & 39.63 & 54.89    & 40.62     & 49.56 \\
\quad-w/o SD \& properties     & 39.66 & 54.95	 & 40.71	 & 49.64 \\
\quad-w/o SD \& recalled pairs & 32.21 & 49.54	 & 34.13	 & 43.34 \\ \bottomrule
\end{tabular}
\end{table}

We conduct our ablation study by removing MsMAAG model of the item properties, recalled pairs and source dropout methods. The results are show in table \ref{tab:remove_source}. Firstly, comparing the there methods conclusion can be drawn that the recalled review-response pairs contribute most to the final performance with an increasemen of 1.63 BLEU score and 1.67 avaraged ROUGE score. The second effective source of information is the properties, and there's only little improvement by using source dropout. However, we add further experiments to simultaneously remove the multi-source prior knowledge and do not apply SD method. This further experiment illustrates the function of source dropout method: when the model is trained without SD, it will degenerate into mostly reliant on the recalled pairs, and when this source of information is removed, we can see a significant drop (7.42 in BELU) in the final performance. Hence, this method is of great use for boosting the robustness of models, especially for some online scenarios where we do not have access to certain source of information.

\begin{figure*}[!h]
  \centering
  \includegraphics[width=\linewidth]{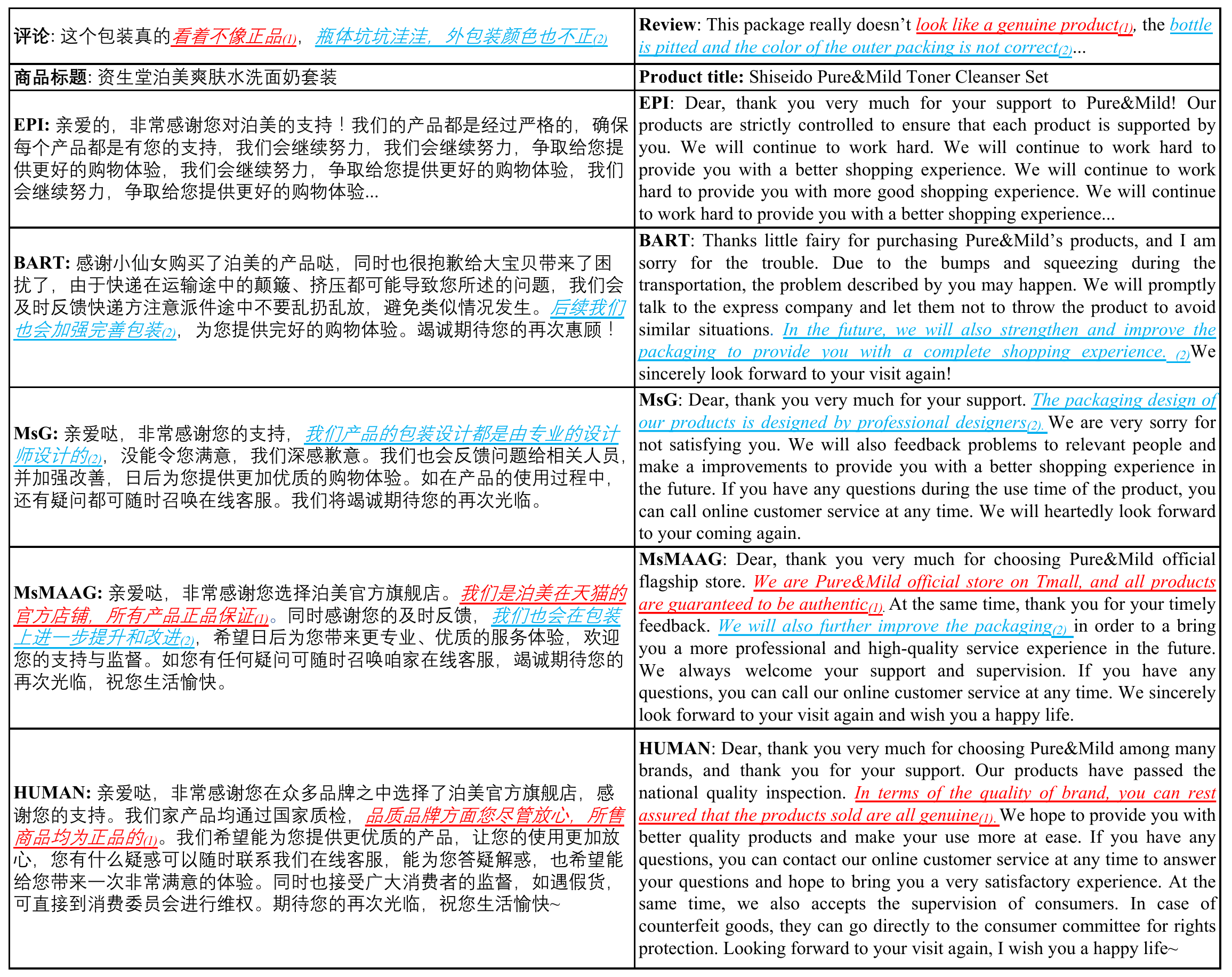}
  \caption{Case study. Different aspects of issues are marked in different color.}\label{fig:response}
\end{figure*}


\subsection{Case Study}

As is illustrate on Figure \ref{fig:response}, we provide an example of generated responses by the models. In this negative review, the consumer mentions two aspects of issues, where one claims that the product seems not authentic, and the other states some problem with the product packing. For EPI model, we notice an unusual repetition of certain phrases. This may due to fact that traditional Bi-GRU network without BART-like pre-training should have much limitation when generating long sentences. BART model and MsG model both respond to the second issue while ignoring the first. The MsMAAG model manages to reply to both of the issues due to the Multi-aspect attentive network, and the store's name "Pure\&Mild official store" is copied into the response. With the claim of being the official store of this brand, the persuasiveness of this response is enhanced, which demonstrates the efficacy of our proposed model.

\section{Deployment}

The proposed method has been successfully deployed for months on taobao.com, one of the largest Chinese e-commerce platforms. With this system, the store's efficiency for dealing with negative reviews have boosted approximately three times on average (measured by the number of resolved reviews per person per day), which shows that our proposed system is practical for real-world use. Meanwhile, the adoption rate of the generated responses is 65\%, indicating that most of them are suitable for use as our system involves specific pre-and-post processing procedure apart from the primary MsMAAG model.

The deployment involves on-line process and off-line process. We use pipeline methods to prepare generated candidate responses in the off-line process, and the stores and sellers can manually adopt appropriate responses with little modification and push them to real on-line reviews. Our whole process is implemented on MaxCompute\footnote{\url{https://www.alibabacloud.com/product/maxcompute}} platform with the support for distributed computing and large-scale data warehousing. Off-line process consists of a sentiment analysis module, a pre-processing module, a ranking module, a MsMAAG module, and a rule-based post-processing module. Sentiment analysis is implemented by a BERT-based classifier. Pre-processing includes data-cleaning, removing common and uninformative reviews and etc. The ranking module involves retrieving recalled historical review-response pairs using ElasticSearch\footnote{\url{https://www.elastic.co/}}. The MsMAAG model is trained on larger-scale business datasets and is deployed on GPU machines. Note that different from our experiments, we pre-process the training dataset by substituting the item-title and store's name with two special tokens [TITLE] and [STORE]. These two special tokens can be generated into the final response, and finally we will fill those tokens with real titles and store names accordingly when representing to the store sellers. This small trick guarantees that we do not produce false information with regard to these facts.

\section{Conclusion}
In this work, we define the task of persuasive responses generation to user reviews in E-commerce scenarios. Such persuasive responses are beneficial for the sellers to deal with negative reviews, which is very useful in practice. In order to solve this problem, we design a Multi-source Multi-aspect Attentive Generation model to firstly encode different sources of information (such as product titles, seller names, product properties and product historical review-response pairs), and then attend to different aspect of issues mentioned in the review, and finally incorporate pointer-generator mechanism to produce response. On two large-scale real-world e-commerce datasets, including one collected by us, we carry our experiments and the results prove the efficacy and the SOTA performance of our model. After the model is deployed online, we observed a three times enhance of efficiency for the store's sellers to deal with negative reviews.

\bibliographystyle{ACM-Reference-Format}
\balance
\bibliography{sample-base}










\end{document}